\pdfoutput=1

\documentclass[11pt]{article}

\usepackage[]{acl}

\usepackage{times}
\usepackage{latexsym}

\usepackage[T1]{fontenc}

\usepackage[utf8]{inputenc}

\usepackage{microtype}

\usepackage{inconsolata}

\usepackage{graphicx}

\usepackage{amsmath}
\usepackage{booktabs}
\usepackage{multirow}

\title{Exploring Multilingual Probing in Large Language Models: \\ A Cross-Language Analysis}

\author{
 \textbf{Daoyang Li\textsuperscript{1,2}},
 \textbf{Haiyan Zhao\textsuperscript{2}},
 \textbf{Qingcheng Zeng\textsuperscript{3}},
 \textbf{Mengnan Du\textsuperscript{2}}
\\
\\
 \textsuperscript{1}University of Southern California,
 \textsuperscript{2}New Jersey Institute of Technology,
 \textsuperscript{3}Northwestern University
\\
 daoyangl@usc.edu, mengnan.du@njit.edu
}

\begin{document}
\maketitle
\begin{abstract}
Probing techniques for large language models (LLMs) have primarily focused on English, overlooking the vast majority of other world's languages. In this paper, we extend these probing methods to a multilingual context, investigating the behaviors of LLMs across diverse languages. We conduct experiments on several open-source LLM models, analyzing probing accuracy, trends across layers, and similarities between probing vectors for multiple languages. Our key findings reveal: (1) a consistent performance gap between high-resource and low-resource languages, with high-resource languages achieving significantly higher probing accuracy; (2) divergent layer-wise accuracy trends, where high-resource languages show substantial improvement in deeper layers similar to English; and (3) higher representational similarities among high-resource languages, with low-resource languages demonstrating lower similarities both among themselves and with high-resource languages. These results highlight significant disparities in LLMs' multilingual capabilities and emphasize the need for improved modeling of low-resource languages.
\end{abstract}

\section{Introduction}
Large language models (LLMs), such as GPT-4~\cite{achiam2023gpt}, Claude 3.5~\cite{anthropic2024claude}, Llama 3~\cite{dubey2024llama}, have demonstrated remarkable progress across a wide range of natural language processing tasks. As these models continue to advance, there is a growing need to understand their internal mechanisms and representations. Probing techniques have emerged as a valuable tool for investigating how LLMs encode and process information, offering insights into their decision-making processes and the nature of their learned representations~\cite{ferrando2024primer,zhao2024opening,zou2023representation}.

However, a significant gap exists in our understanding of LLMs' multilingual capabilities. While extensive probing research has been conducted on English language representations, there are approximately 7,000 languages spoken worldwide, many of which remain understudied in the context of LLMs. This lack of comprehensive multilingual analysis limits our understanding of how LLMs perform across diverse linguistic contexts, particularly for low-resource languages that are often underrepresented in model training data and evaluations.

To address this research gap, we propose a multilingual probing approach to investigate the behavior of LLMs across a diverse set of 16 languages, including both high-resource and low-resource languages. Our study extends probing techniques from English to a multilingual context, examining how LLMs perform in factual knowledge and sentiment classification tasks across different languages. Our key findings reveal that: (1) high-resource languages consistently achieve higher probing accuracy compared to low-resource languages; (2) high-resource languages exhibit similar trends to English across model layers, with accuracy improving significantly in deeper layers, while low-resource languages show relatively stable or only slightly improving accuracy; and (3) there are high similarities between probing vectors of high-resource languages, whereas low-resource languages demonstrate lower similarities both among themselves and with high-resource languages.

\section{Probing Method}
\subsection{LLM Internal Representation}

We study decoder-only LLMs, where each layer of a model consists of both multi-head attention blocks (MHA) and feed-forward networks (FFNs). In this work, we utilize frozen pretrained language models. Layers are indexed with $\ell \in L$, where $L$ denotes the set of all layers in a model.
For each layer, the computation starts and ends with a residual stream. The MHA first reads from the residual stream and performs computation, then adds its output back to the residual stream. The updated vector in the residual stream is then passed through MLPs to generate the output of the layer:
\begin{equation}
\small
h_i^{\ell+1} = h_i^\ell + \text{MLP}^\ell \left(h_i^\ell + \text{Att}^\ell \left(h_i^\ell\right)\right),
\end{equation}
where $h_i^\ell$ represents the hidden state of the $i$-th token in the input sequence at layer $\ell$.
We focus on the output representation space of each layer, particularly the residual stream at the end of each layer. We use the representation of the last token to represent the entire input sequence, as it is generally believed to integrate information from all previous tokens. This representation is denoted as $h^\ell$, which will be simplified to $h$ in the following section.

\subsection{Linear Classifier Probing}
In our analysis, we employed linear classifier probing \cite{ju2024how, jin2024exploringconceptdepthlarge} to explore internal representations across various layers of LLMs.
We extracted hidden states from the residual stream of each layer using two types of inputs (i.e., positive and negative) and utilized these representations to train a logistic regression model. By evaluating the performance of trained classifier, we are able to assess how well the hidden states at different layers encoded information relevant to answering factual questions or handling sentiment classification tasks. This approach provides valuable insights into the nature of the representations learned within the model.

To perform the probing, we employed a linear classifier approach. We define $h \in {R}^{n \times d_{\text{model}}}$ as the set of hidden features extracted from the LLM, where $n$ is the number of samples and $d_{\text{model}}$ represents the dimensionality of the hidden layer. The internal representation of each sample in a specific layer is denoted by $h^{(i)} \in {R}^{1 \times d_{\text{model}}}$. We utilize binary classification, assigning labels $y^{(i)} \in \{0, 1\}$. The objective function for our logistic regression classifier, incorporating L2 regularization, is formulated as:
\begin{equation}
\small
J(\theta) = -\frac{1}{n} \sum_{i=1}^{n} L(h^{(i)}, y^{(i)}; \theta) + \frac{\lambda}{2n} \|\theta\|_2^2,
\end{equation}
where $L(.)$ represents the cross-entropy loss:
\begin{equation}
\small
L = y^{(i)} \log(\sigma(\theta^T h^{(i)})) + (1-y^{(i)}) \log(1-\sigma(\theta^T h^{(i)})),
\end{equation}
where $\theta$ denotes the model parameters, $\lambda$ is the regularization coefficient, and $\sigma(\cdot)$ represents the sigmoid activation function. By evaluating the accuracy of this classifier on the test set, we can evaluate the LLM's performance and gain insights into its internal representations across different languages and layers.

\begin{figure*}[h]
    \centering
    \includegraphics[width=1\textwidth]{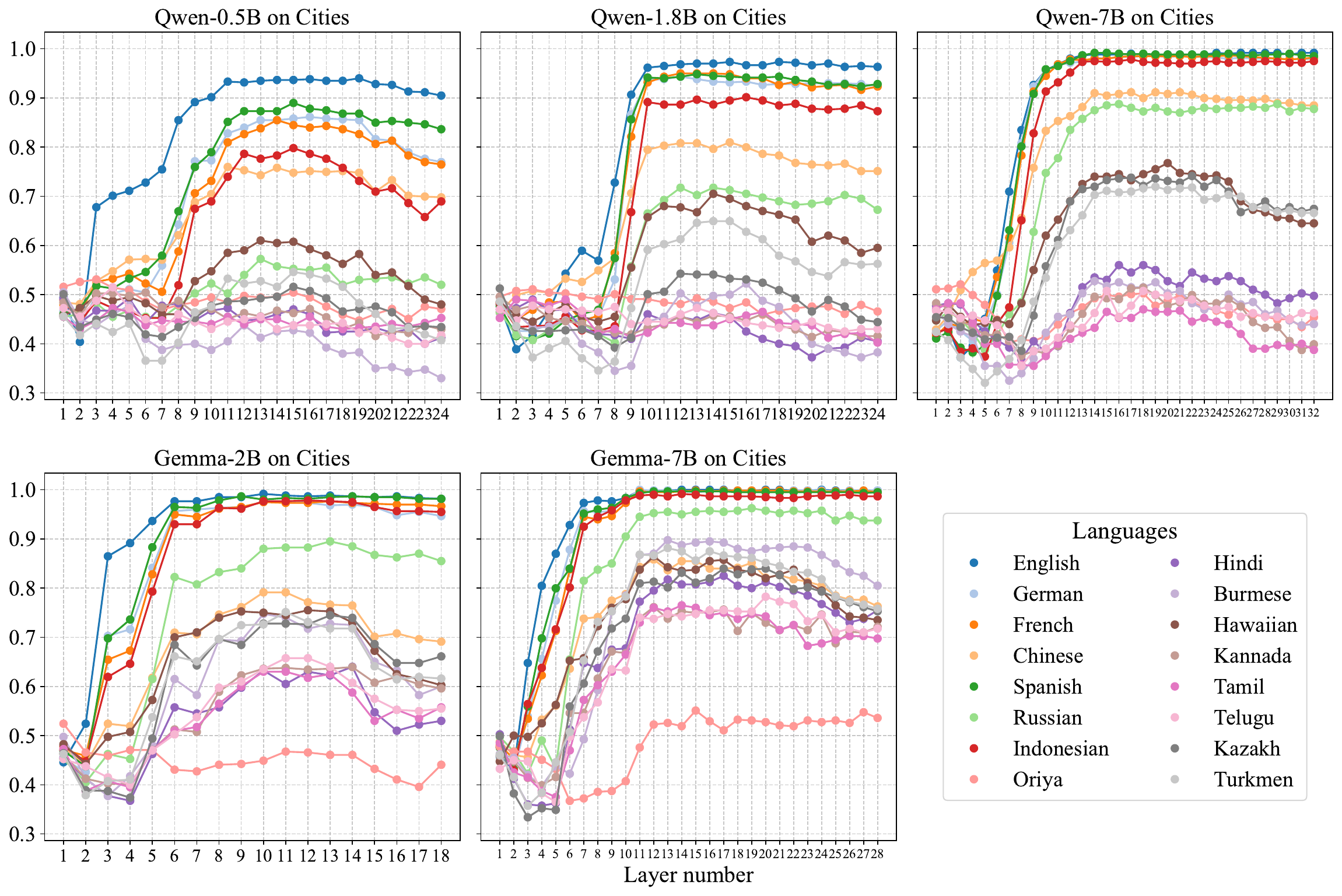}
    \caption{Layer-wise probing accuracy of 5 open-source LLMs across 16 languages.}
    \label{fig:Qwen0.5_cities}
\end{figure*}

\section{Experiment}
In this section, we conduct comprehensive probing experiments to investigate how language models process different languages. Our analysis focuses on two key aspects: comparing accuracy across different model layers and examining correlations between probing vectors of various languages. Through these experiments, we seek to answer three fundamental research questions (\textbf{RQ}s):
\begin{itemize}
\setlength\itemsep{-0.3em}
\item \textbf{RQ1} - Do other languages achieve comparable probing accuracy to English?
\item \textbf{RQ2} - Do other languages exhibit similar layer-wise behavioral patterns to English?
\item \textbf{RQ3} - What are the similarities between probing vectors across different languages?
\end{itemize}

\begin{table}[t]
\centering
\small
\begin{tabular}{lccc}
\hline
\textbf{Model} & \textbf{Layer} & \textbf{Representation Dimension} \\
\hline
Qwen-0.5B & 24 & 1024 \\
Qwen-1.8B & 24 & 2048 \\
Qwen-7B & 32 & 4096 \\
Gemma-2B & 18 & 2048 \\
Gemma-7B & 28 & 3072 \\
\hline
\end{tabular}
\caption{Model and corresponding layers.}
\label{tab:modelandlayer}
\end{table}
\subsection{Experiment Settings}
In this section, we introduce the overall experimental settings of this papers.

\noindent\textbf{Models:}\,
We evaluated the performance and internal representations across various languages using two open-source LLM families: Qwen \cite{bai2023qwen} and Gemma \cite{gemma2024}. The Qwen-1.5 architecture comprises 24 layers for smaller variants (0.5B \& 1.8B) and 32 layers for the larger variant (7B), while the Gemma architecture features 18 layers for the 2B model and 28 layers for the 7B model. The representation vector dimensions vary across models: 1024 for Qwen-0.5B, 2048 for both Qwen-1.8B and Gemma-2B, 3072 for Gemma-7B, and 4096 for Qwen-7B. Table~\ref{tab:modelandlayer} provides a comprehensive overview of the models and their corresponding layer configurations.

\noindent\textbf{Datasets:}\,
In the following experiments, we utilized a truthful dataset: \emph{Cities} \cite{marks2023geometry}, and a sentiment dataset: \emph{Opinion} \cite{deceptive_opinion_spam_corpus}. Cities contains 1496 samples, and Opinion contains 1000 samples.
\begin{itemize}\setlength\itemsep{-0.3em}

\item \emph{Cities} \cite{marks2023geometry}: consists of statements about the location of cities from worldwide and their veracity labels  (e.g., The city of Lyon is in France, which is true).

\item \emph{Opinion} \cite{deceptive_opinion_spam_corpus}: consists of opinions of 20 famous hotels. It contains the hotel's name, opinion's polarity, and its source.
\end{itemize}
Our dataset encompasses 16 languages: English, German, French, Chinese, Spanish, Russian, Indonesian, Oriya, Hindi, Burmese, Hawaiian, Kannada, Tamil, Telugu, Kazakh, Turkmen. We categorized English, German, French, Chinese, Spanish, Russian, and Indonesian as high-resource languages, and rest of them as low-resource languages based on the volume of available digital content and linguistic resources. The original language of our two datasets are English, and we used Google Translate within deep-translator python library \cite{deeptranslator} to translate them into other 15 languages, as Google Translate supports translation between over 100 languages, and achieves high accuracy compared to other translation tools.

\begin{table*}[t]
\centering
\caption{Probing accuracy of various LLMs across different languages on the Cities dataset.}
\setlength{\tabcolsep}{1.6pt}  
\scalebox{0.65}{  
\begin{tabular}{lccccccccccccccccc}
\toprule
\multirow{2}{*}{\textbf{Model}} & \multicolumn{7}{c}{\textbf{High-Resource Languages}} & \multicolumn{9}{c}{\textbf{Low-Resource Languages}} \\
\cmidrule(lr){2-8} \cmidrule(lr){9-17}
& {English} & {German} & {French} & { Chinese} & { Spanish} & { Russian} & { Indonesian} & { Oriya} & { Hindi} & { Burmese} & { Hawaiian} & { Kannada} & { Tamil} & { Telugu} & { Kazakh} & { Turkmen} \\
\midrule
Gemma-2B & 0.98 & 0.95 & 0.97 & 0.69 & 0.98 & 0.87 & 0.95 & 0.44 & 0.53 & 0.60 & 0.60 & 0.60 & 0.56 & 0.56 & 0.66 & 0.62 \\
Gemma-7B & 0.99 & 0.99 & 0.99 & 0.76 & 0.99 & 0.93 & 0.99 & 0.54 & 0.76 & 0.81 & 0.74 & 0.72 & 0.70 & 0.72 & 0.75 & 0.76 \\
Qwen-0.5B & 0.90 & 0.77 & 0.76 & 0.70 & 0.84 & 0.52 & 0.69 & 0.47 & 0.41 & 0.33 & 0.48 & 0.43 & 0.45 & 0.42 & 0.43 & 0.41 \\
Qwen-1.8B & 0.96 & 0.92 & 0.92 & 0.75 & 0.93 & 0.67 & 0.87 & 0.47 & 0.41 & 0.37 & 0.60 & 0.42 & 0.40 & 0.43 & 0.44 & 0.56 \\
Qwen-7B & 0.99 & 0.98 & 0.98 & 0.88 & 0.98 & 0.88 & 0.97 & 0.45 & 0.50 & 0.44 & 0.65 & 0.40 & 0.39 & 0.46 & 0.67 & 0.67 \\
\bottomrule
\end{tabular}
}
\label{tab:finalaccuracy}
\end{table*}

\vspace{5pt}
\noindent\textbf{Implementation Details:}\, To evaluate the performance of LLMs on each language, we use the template for the Cities dataset in English as "\textit{Judge the statement is Positive or Negative. <Statement>}". The prompts of other languages utilize the same template translated by Google Translate. This allows us to prevent any context differences regarding the prompt design. We present the full set of prompt templates for all 16 languages in Figure~\ref{fig:prompt} at the Appendix. We applied probing techniques to assess the information encoded within each layer of these models. For our probing analysis, we selected linear classifier probing for our experiments. Each dataset is divided into a training and a test set with an 8:2 ratio, and we adhered to the standard procedure for probing classifiers in LLMs, extracting feature representations from the final hidden states at each layer of the LLMs to serve as input to the probing classifier. The linear weight parameter $\theta$ of the logistic regression classifier is regarded as the probing vector for each language and layer.

\subsection{Multilingual Accuracy}
In this section, we explored (1) whether other languages besides English have the same probing accuracy as English and (2) whether they follow the same trend as English in different layers.

We present results on multilingual accuracy across our five evaluated models (Qwen-0.5B, Qwen-1.8B, Qwen-7B, Gemma-2B, Gemma-7B) on the cities and Opinion datasets. In Figure~\ref{fig:Qwen0.5_cities} and Table~\ref{tab:finalaccuracy}, we show the results of layer-wise probing accuracy on the Cities dataset. The results of Opinion dataset are included in Figure~\ref{fig:opinion_acc_appendix} in the Appendix. These results visualize how probing accuracy changes across model layers for all 16 languages. Based on these results, our analysis lead to two general observations as follows:

\begin{itemize}\setlength\itemsep{-0.3em}
\item
\emph{High-resource languages show higher accuracy, while low-resource languages have comparatively lower accuracy.} We conducted experiments using Cities and Opinion datasets, exploring the binary classification problem in 16 selected languages. Table~\ref{tab:finalaccuracy} shows that in Cities dataset, high-resource languages such as French and German achieve at least 70\% accuracy, even reaching over 90\% accuracy for some models, while low-resource languages like Oriya and Hindi only achieve about 40\% accuracy in the final layer.

\item
\emph{High-resource languages follow similar trends to English, where accuracy significantly improves as the layers deepen. Low-resource languages maintain relatively stable probing accuracy or show only slight improvements.} Figure~\ref{fig:Qwen0.5_cities} shows that as model layers go deeper, English, French, and other high-resource languages could reach highest accuracy at the 11th layer. However, the probing accuracies of the low-resource languages have not improved significantly.
\end{itemize}

\begin{figure*}[t]
    \centering
    \includegraphics[width=1\textwidth]{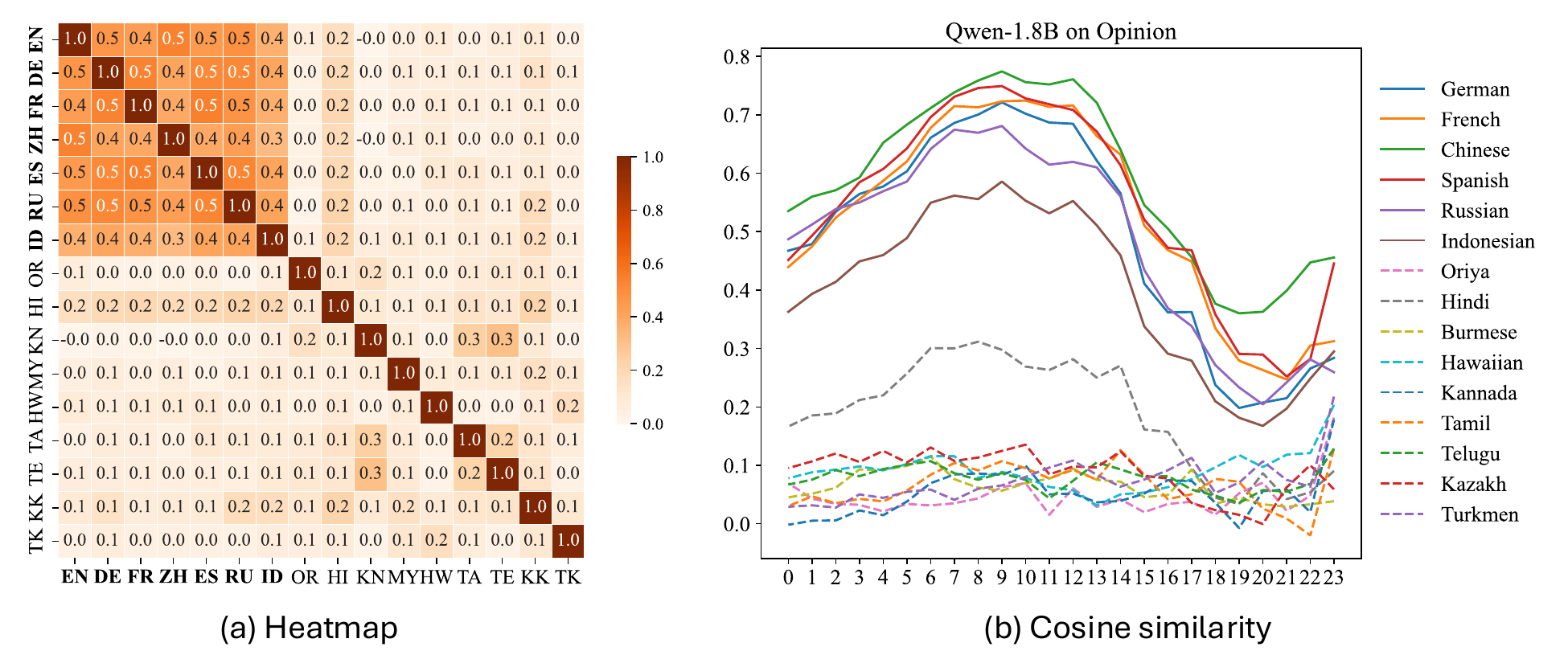}
    \caption{(a) Heatmap of the similarities of probing vectors correlation across languages; (b) Cosine similarity of probing vectors with English. (Model: Qwen-1.8B, Dataset: Opinion).}
    \label{fig:Qwen1.8_opinion}
\end{figure*}

\subsection{Similarity Correlation of Probing Vectors }

In this section, we conducted similarity analysis on probing vectors $\theta$ across languages using two visualization approaches: 
\begin{itemize}\setlength\itemsep{-0.3em}
\item \textbf{Correlation Heatmaps}: These visualize the pairwise similarities between probing vectors of all 16 languages. These highlight clustering patterns and resource-level disparities.
\item \textbf{Layer-wise Similarity Plots}: They measure cosine similarity between each language's probing vector and English's across model layers, revealing representation dynamics.
\end{itemize}
For demonstration, Figure~\ref{fig:Qwen1.8_opinion} shows results from the Qwen-1.8B model and Opinion dataset. In the Appendix, we extend this analysis to all five models (Qwen-0.5B, Qwen-1.8B, Qwen-7B, Gemma-2B, Gemma-7B) and both datasets (Opinion, Cities) through Figure~\ref{fig:heatmap-similarities} and Figure~\ref{fig:cosine-similarity}. Our analysis reveals the following three key patterns:
\begin{itemize}\setlength\itemsep{-0.3em}
\item The probing vectors of high-resource languages (English, German, French, Chinese, Spanish, Russian) demonstrate strong correlations with each other, as evidenced by the darker clusters in the heatmaps and consistently higher similarity curves in the trajectory plots. For instance, in the Qwen-1.8B Opinion task, German and French probing vectors maintain correlations above 0.6 with English across most layers. In contrast, low-resource languages show notably weaker correlations, both among themselves and with high-resource languages. This pattern is visible in the bright regions of the heatmaps for languages like Tamil, Telugu, and Oriya, with similarity scores typically remaining below 0.3 across all layers.

\item
The evolution of similarities across model layers reveals further insights into these representational differences. High-resource languages exhibit dynamic similarity patterns with English, often peaking in middle layers before slightly decreasing, while low-resource languages maintain relatively stable, low similarity levels throughout the model layers. These patterns persist across different model sizes and architectures in both the Qwen and Gemma families, and remain consistent across the Opinion and Cities datasets.
\end{itemize}

\section{Related Work}
In this section, we review two lines of research that are most relevant to ours.

\noindent
\textbf{Multilingual Abilities of LLMs.} The multilingual capabilities of LLMs have garnered increasing attention from researchers \cite{ali2024surveylargelanguagemodels, jayakody2024performancerecentlargelanguage}. Recent studies have investigated the consistency of factual knowledge across different languages in multilingual pretrained language models (PLMs)~\cite{fierro2022factual,qi2023cross}. Additionally, significant efforts have been directed towards enhancing the representation of low-resource languages \cite{abadji-etal-2022-towards, imanigooghari-etal-2023-glot500, li2024quantifying}. These investigations demonstrate that LLMs still possess considerable untapped potential in multilingual capabilities.

\vspace{0.3cm}
\noindent
\textbf{Probing Representations in LLMs.} Probing is a popular method to investigate the internal representations for LLMs in recent days, which is widely used in LLM interpretability studies \cite{alain2018understandingintermediatelayersusing, taktasheva-etal-2021-shaking, pimentel2020informationtheoreticprobinglinguisticstructure, ferrando2024primerinnerworkingstransformerbased,wendler2024llamas}. Previous work demonstrate that different layers typically acquired different information \cite{jin2024exploringconceptdepthlarge, ju2024largelanguagemodelsencode}. Various works using probing technique to assess how they encode linguistic features \cite{liu2023cognitivedissonancelanguagemodel, marks2024geometrytruthemergentlinear}. 

Our study employs probing techniques to examine LLMs' performance and internal representations across different languages. The most closely related work is the Language Ranker~\cite{li2024quantifying}, which uses cosine similarity between a language's representation and English as a baseline. In contrast, our method utilizes linear classifier probing to evaluate performance across languages. This approach allows us to directly assess the model's ability to extract language-specific information, providing a more detailed view of LLMs' multilingual capabilities.

\section{Conclusions and Future Work}

In this work, our multilingual probing experiments on LLMs reveal significant disparities in performance and representational qualities across languages, suggesting potential limitations in how these models learn linguistic concepts. Specifically, high-resource languages consistently achieve higher probing accuracy and exhibit similar trends to English, with accuracy improving significantly in deeper layers. We also observe high similarities between probing vectors of high-resource languages, while low-resource languages demonstrate lower similarities both among themselves and with high-resource languages. These findings not only indicate the current limitations of LLMs in handling low-resource languages, but also suggest that these models may not be learning deeper linguistic concepts effectively across all languages.

In future, we plan to conduct research to address these gaps by developing more equitable and effective multilingual language models that can better capture universal linguistic concepts. Besides, we plan to extend this research to multimodal models that incorporate visual and textual information.

\clearpage

\section*{Limitations}
In this work, we use machine translation to generate the prompt templates and question sentences from English to other languages, which may introduce noise. We only experiment with five open-source LLMs and two datasets. In the future, we would like to expand these findings with other datasets and models to confirm how well the LLMs' performance and representations generalize in these settings. Additionally, we just utilized linear classifier probing to do the experiments. We plan to explore more sophisticated probing methods beyond linear classifiers, which could offer deeper insights into the nature of linguistic representations within LLMs. 

\section*{Acknowledgment}
The work is in part supported by NSF \#2310261. The views and conclusions in this paper are those of the authors and should not be interpreted as representing any funding agencies.

\bibliography{reference}

\clearpage
\appendix

\label{sec:appendix}

\begin{figure*}[h]
    \centering
    \includegraphics[width=1\textwidth]{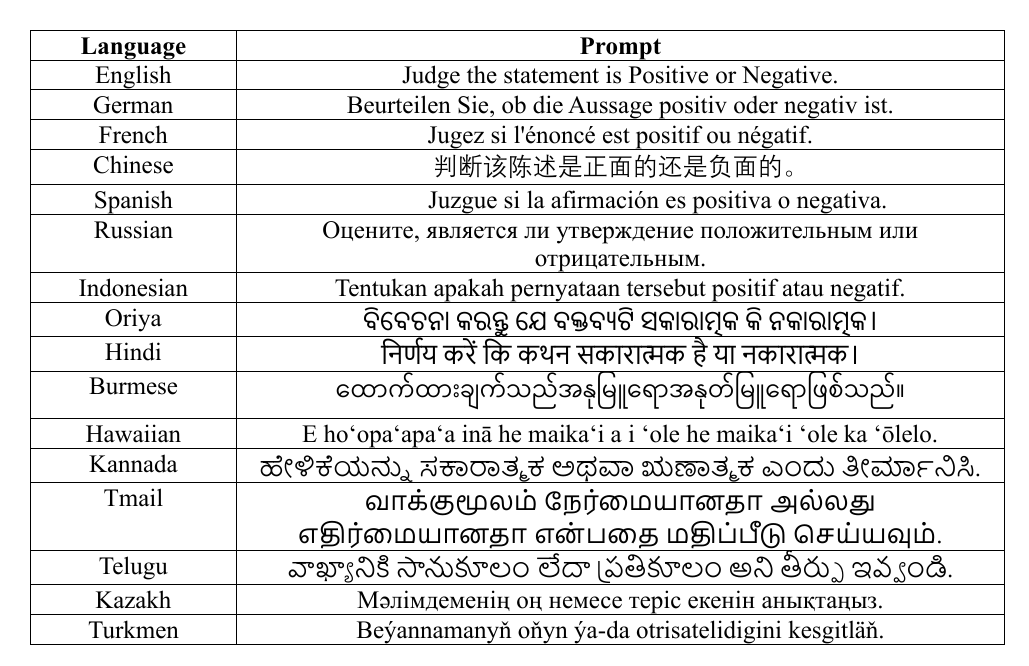}
    \caption{Prompt templates of all languages used in experiments.}
    \label{fig:prompt}
\end{figure*}

\begin{figure*}[h]
    \centering
    \includegraphics[width=1\textwidth]{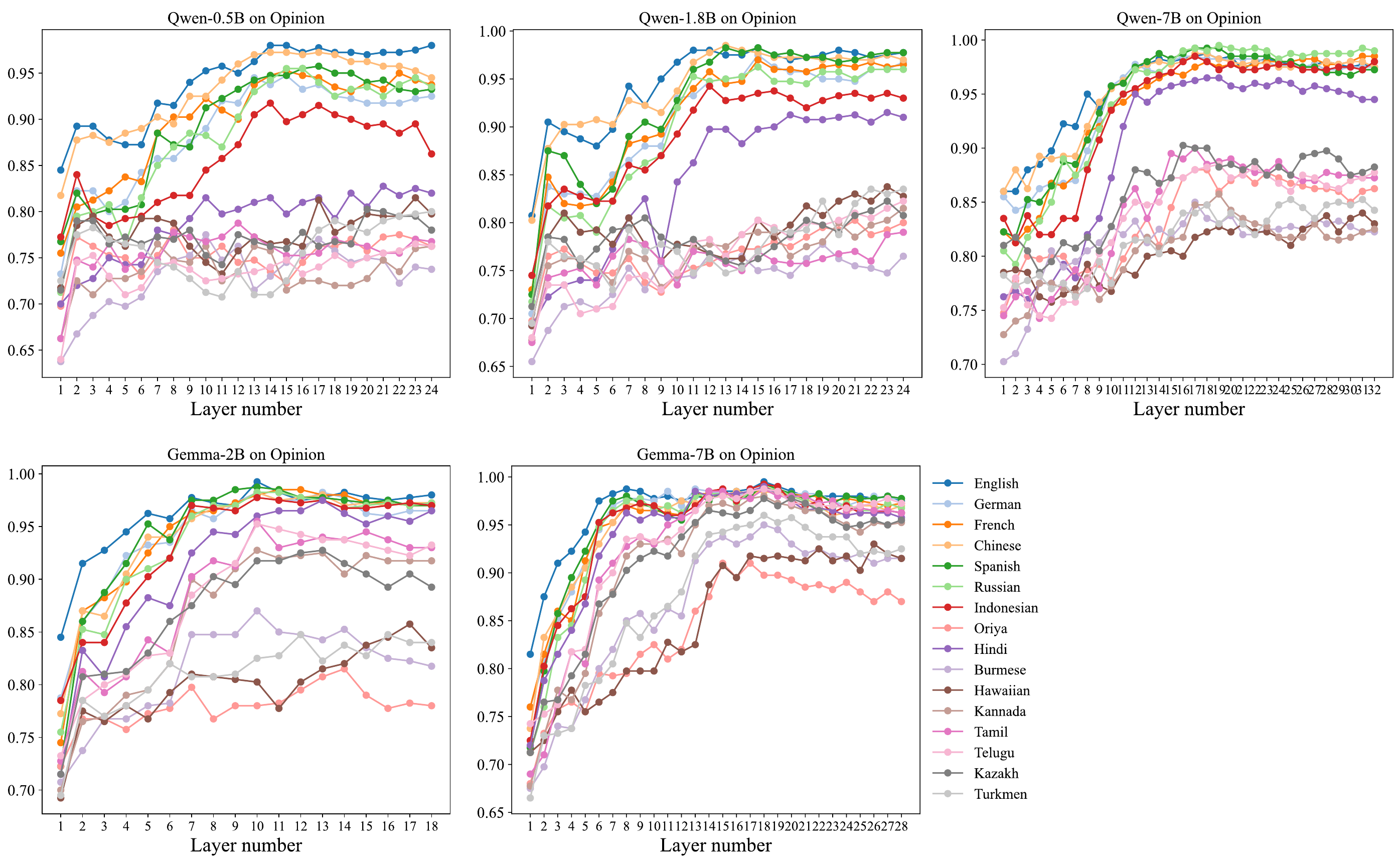}
    \caption{Additional results for multilingual accuracy of Qwen and Gemma Series Model on the Opinion Dataset}
\label{fig:opinion_acc_appendix}
\end{figure*}

\begin{figure*}
    \centering
    \includegraphics[width=1\textwidth]{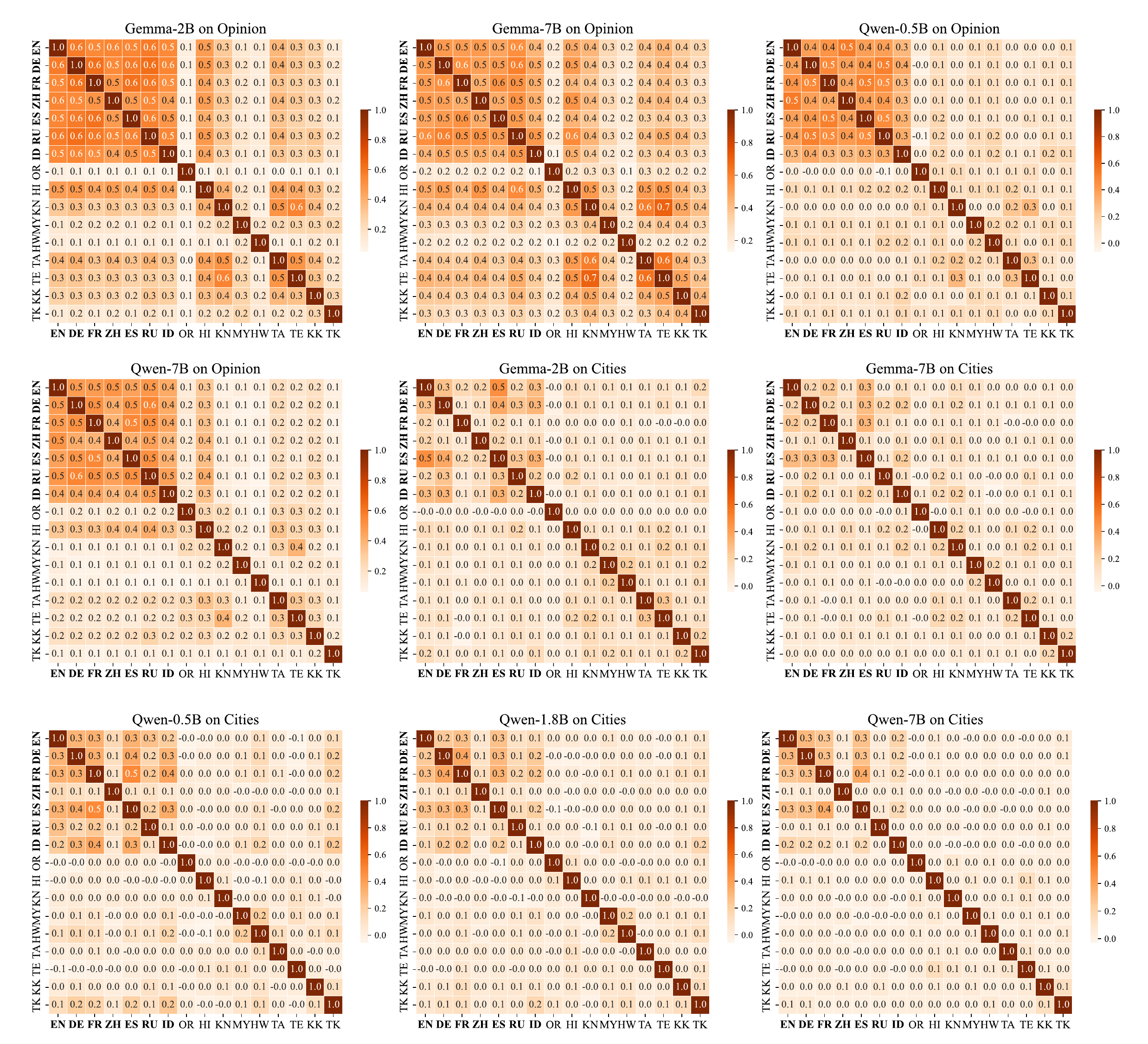}
    \caption{Heatmap of the similarities of probing vectors correlation across languages.}
    \label{fig:heatmap-similarities}
\end{figure*}

\begin{figure*}
    \centering
    \includegraphics[width=1\textwidth]{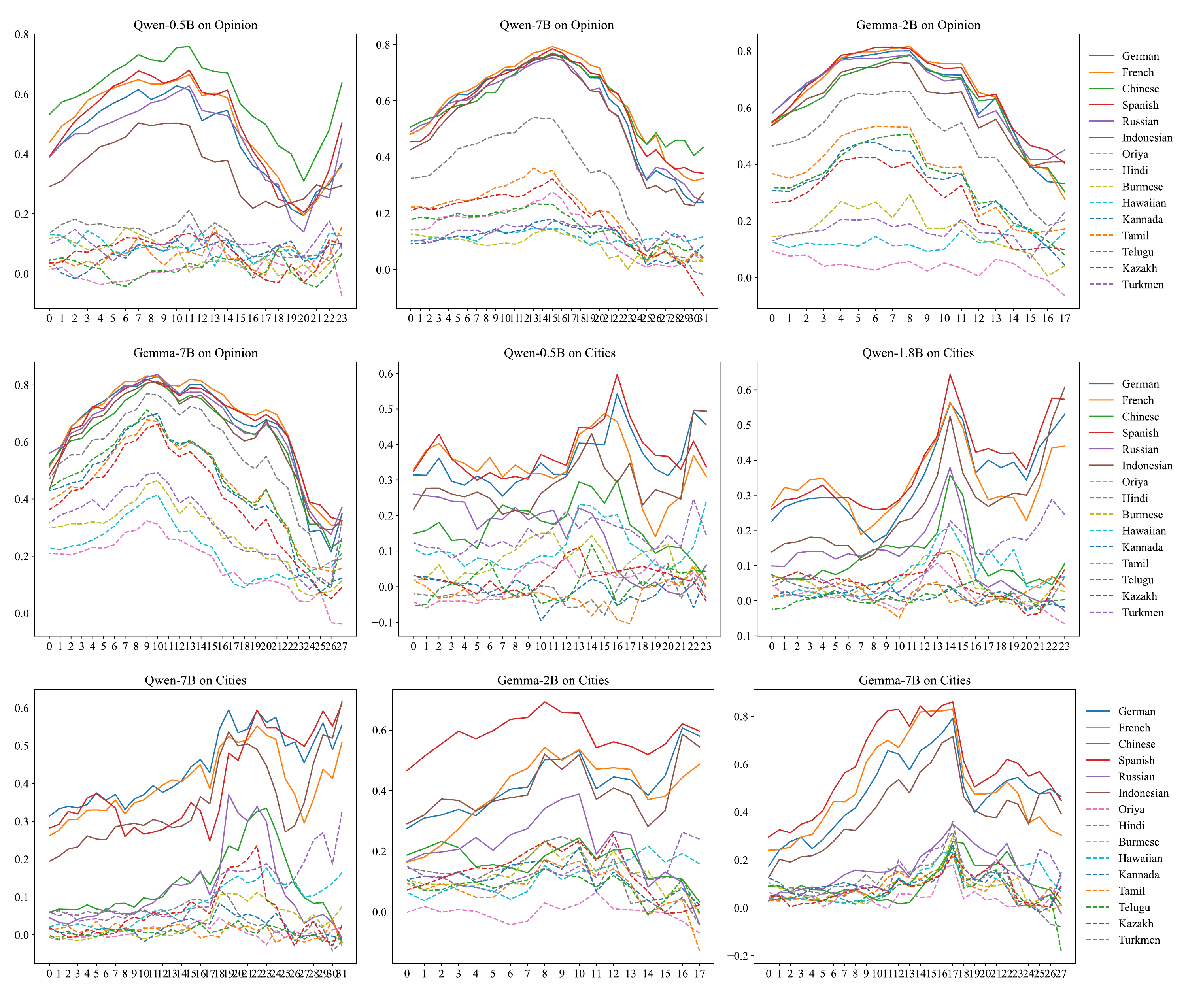}
    \caption{Cosine similarity of probing vectors with English across different language models (Qwen-0.5B, Qwen-7B, Gemma-2B, and Gemma-7B) and tasks (Opinion and Cities).}
    \label{fig:cosine-similarity}
\end{figure*}

\end{document}